\documentclass[11pt]{article}

\usepackage[preprint]{acl}

\usepackage{times}
\usepackage{latexsym}

\usepackage[T1]{fontenc}

\usepackage[utf8]{inputenc}

\usepackage{microtype}

\usepackage{inconsolata}

\usepackage{graphicx}

\usepackage{amsmath}
\usepackage{amssymb}        
\usepackage{amsthm} 
\usepackage{algorithm}      
\usepackage{algorithmic}    
\usepackage{booktabs}       
\usepackage[table]{xcolor}  
\usepackage{subcaption}
\usepackage{siunitx, booktabs, array}
\usepackage{hyperref}
\usepackage{fontawesome5}

\definecolor{linkpink}{RGB}{219, 48, 122}

\title{Joint Training of Multi-Token Prediction in Reinforcement \\Learning via Optimal Coefficient Calibration}

\author{Zili Wang$^{1,2}$\thanks{Work done as an intern at Meituan.}, 
  Jiajun Chai$^{3}$,  
  Lin Chen$^{1,2}$,
  Xiaohan Wang$^{3}$, 
  \\
  \bf Shiming Xiang$^{1,2}$, 
  Guojun Yin$^{3}$
  \\
  $^1$ School of Artificial Intelligence, University of Chinese Academy of Sciences, China\\
  $^2$ MAIS, Institute of Automation, Chinese Academy of Sciences, China\quad
  $^3$ Meituan \\\\
  \faGithub\ Code:
\href{https://github.com/MarkXCloud/RL-MTP}
{\texttt{https://github.com/MarkXCloud/RL-MTP}}
}

\begin{document}
\maketitle
\begin{abstract}
    Reinforcement Learning from Verifiable Rewards (RLVR) has emerged as the standard paradigm for improving reasoning capability of large language models, while Multi-Token Prediction (MTP) has been a widely adopted module in pretraining. Combining them is a natural approach, yet current RL practices detach MTP gradients because joint training degrades the performance. We revisit this failure from an optimization perspective. We show that the per-step effect of MTP on the RL objective can be decomposed into two terms: a first-order correlation and a second-order perturbation penalty. This decomposition unifies three MTP training regimes: Detach, Cross-Entropy loss, and Policy loss, and explains why each succeeds or fails. Further analysis of policy loss reveals that, although it aligns with intuition, performance still degrades: the correlation term decays while the quadratic penalty persists. Guided by the analysis, we propose \textbf{Optimal Coefficient Calibration (OCC)}, an adaptive scheme that tracks the optimal coefficient online via a log-probability proxy at negligible cost. Across six competition-level mathematical reasoning benchmarks, OCC consistently matches or exceeds the detach baseline, delivering improved joint MTP-RL training performance.
\end{abstract}
\vspace{-8pt}
\section{Introduction}
\label{sec:introduction}

Reinforcement Learning from Verifiable Rewards (RLVR) has become the standard paradigm for enhancing the reasoning capability of Large Language Models (LLMs)~\cite{schulman2017ppo,shao2024deepseekmath,yu2025dapo} during post-training. Concurrently, Multi-Token Prediction (MTP)~\cite{gloeckle2024metamtp,stern2018blockwise,liu2024deepseekv3}, which is trained to predict multiple future tokens after the main model, has demonstrated substantial benefits during pretraining, including capturing multi-step representations, improving downstream accuracy and serving as a native draft model for speculative decoding~\cite{leviathan2023fast,chen2023accelerating}. Given these strengths, an intuitive approach is to \emph{jointly} optimize MTP and RL objectives during post-training, with the expectation that the model may benefit from multi-token supervision while acquiring reasoning abilities.
\begin{figure*}[t]
\centering
\begin{subfigure}[b]{0.30\linewidth}
    \centering
    \includegraphics[width=\linewidth]{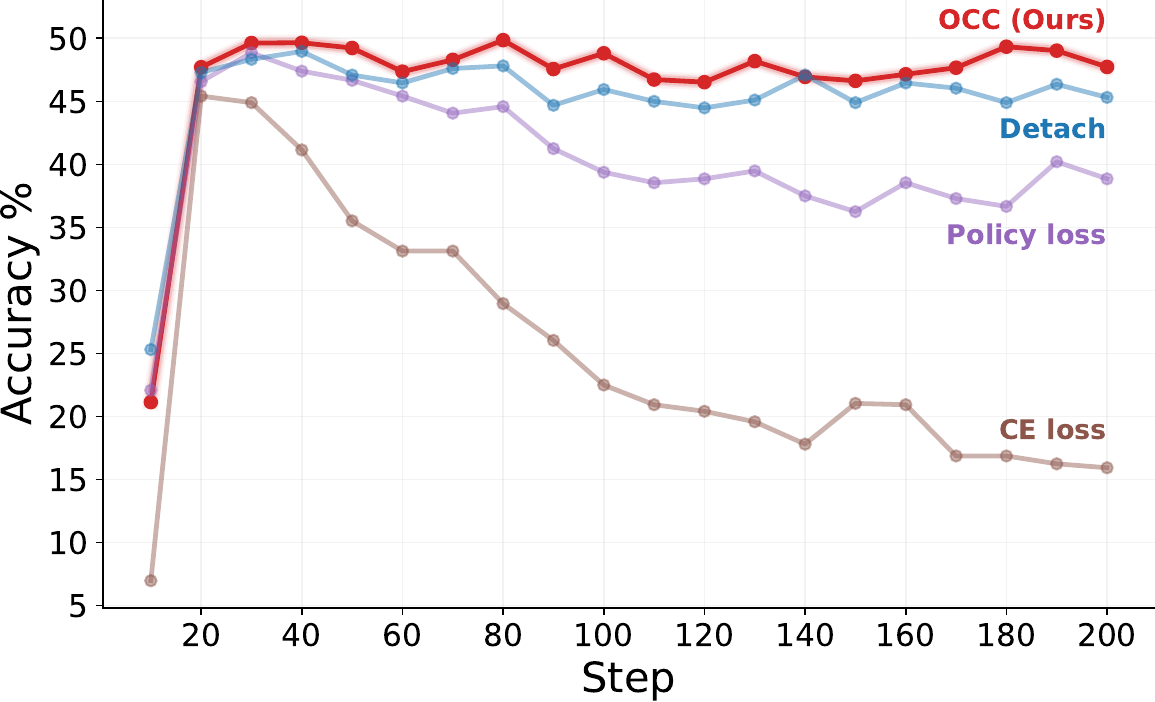}
    \label{fig:aime24acc_sub}
\end{subfigure}
\hfill
\begin{subfigure}[b]{0.69\linewidth}
    \centering
    \includegraphics[width=\linewidth]{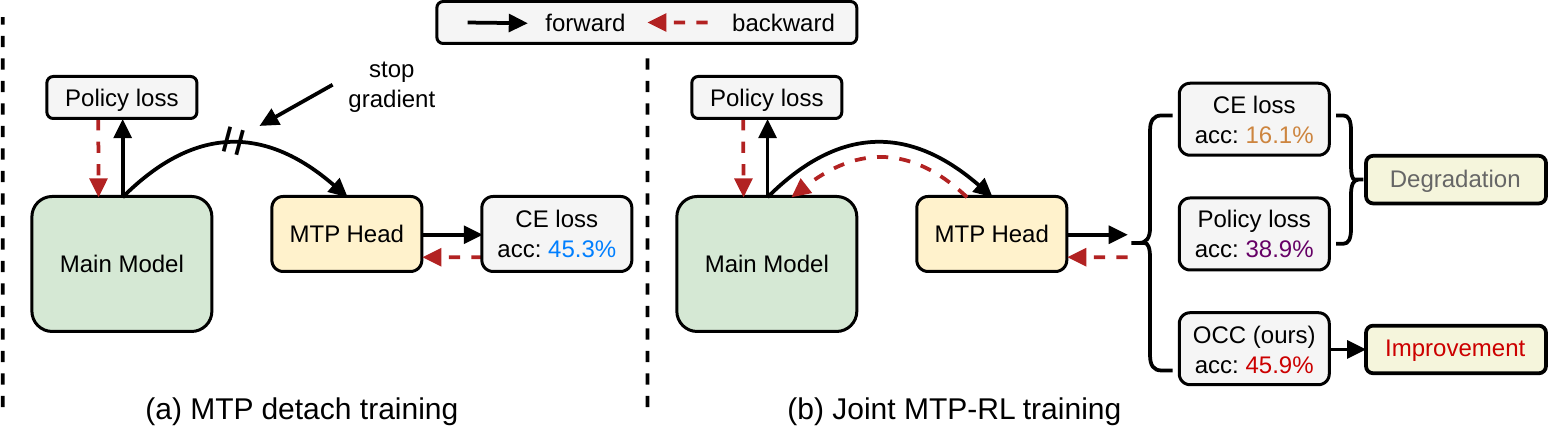}
    \label{fig:main_sub}
\end{subfigure}
\vspace{-30pt}
\caption{\textbf{Left:} Accuracy on AIME24. CE loss shows great degradation, while policy loss initially surpasses but degrades as training progresses. In contrast, OCC sustains gains throughout training and consistently outperforms. \textbf{Middle \& Right:} Illustrations of the MTP training regimes. Detach training stops the gradient from MTP into the main model. Joint MTP-RL training allows this gradient to flow back into the main model, enabling joint updates.}
\label{fig:aime24acc}
\end{figure*}

However, \emph{previous joint MTP-RL training shows significant performance degradation on the main model}, as shown in Figure~\ref{fig:aime24acc}. Therefore, previous practices choose to detach MTP head as a compromise. Widely adopted RL frameworks (e.g., veRL~\cite{sheng2025verl}, slime~\cite{slime}) explicitly report the degradation and recommend gradient detachment~\cite{verl_mtp_docs,zhao2024slimetutorial}. This has pushed recent prominent models toward similar choices. GLM-5~\cite{glm5team2026glm5} and Composer-2~\cite{research2026composer2} isolate MTP head during RL training. Nemotron-3 Super~\cite{nvidia_nemotron_3_super} freezes the main model after RL convergence and fine-tunes MTP head separately. When MTP is detached, its parameters can only adapt to the main model through an independent post-hoc fine-tuning procedure, sacrificing the potential gains from multi-token supervision.

This leads to a natural question: \emph{Why does joint MTP-RL training fail, and can we design a principled strategy to make it work?}

In this paper, we first analyze the joint MTP-RL training from an optimization perspective. We decompose the per-step effect of MTP into: (a) a \emph{first-order correlation term}, determined by the directional agreement between RL and MTP gradients, and (b) a \emph{second-order penalty term}, reflecting the perturbation introduced by MTP gradients. Performance improvement depends on whether the correlation term outweighs the penalty term. 

This decomposition unifies three MTP training regimes: (1)~\textbf{Detach} isolates the MTP gradient from the main model, same as training main model alone. (2)~\textbf{Cross-Entropy (CE) Loss} treats all samples equally with cross-entropy loss, whereas RL up-weights high-reward samples and suppresses low-reward ones; the unrelated objective yields a weak correlation term that cannot outweigh the perturbation. (3)~\textbf{Policy Loss} uses the same RL objective as the main model, so gradients are initially well aligned and the correlation term dominates; however, as the gradient approaches a flat region~\cite{li2018visualizing}, this alignment decays while the perturbation persists, causing a rise-then-fall performance curve. A fixed MTP coefficient cannot track this phase transition, motivating an adaptive coefficient to calibrate the drift throughout training.

Based on this analysis, we propose \textbf{Optimal Coefficient Calibration (OCC)}, an adaptive procedure that calibrates the optimal MTP coefficient during training. OCC starts from the closed-form optimum implied by our analysis. To avoid computing full-model gradient in large-scale distributed training system, we use a log-probability proxy derived from the log probability change under a small-step approximation~\cite{li2026otb,ma2026fipo}. This proxy allows OCC to track the theoretically preferred coefficient online with negligible computational overhead.

We evaluate on multiple competition-level mathematical reasoning benchmarks with different models and algorithms compared with the three training regimes. Extensive results show that cross-entropy loss underperforms the Detach. Policy loss exhibits a rise-then-fall trend. In contrast, our OCC consistently matches or exceeds the Detach across all benchmarks, demonstrating stable improvements across tasks.

Our contributions are summarized as follows:
\begin{enumerate}
    \item We provide a theoretical analysis of joint MTP-RL training. This analysis unifies three MTP training regimes and explains why each succeeds or fails.
    \item We propose \textbf{Optimal Coefficient Calibration (OCC)} that adaptively calibrates the MTP coefficient online using a log-probability proxy, requiring negligible computational overhead and no full-model gradient computation.
    \item We conduct extensive experiments on multiple competition-level mathematical reasoning benchmarks across different models and algorithms. Results demonstrate that OCC consistently matches or exceeds the Detach, demonstrating stable improvements across tasks.
\end{enumerate}
\section{Related Work}
\label{sec:related_work}
\paragraph{Multi-Token Prediction.}
The concept of predicting multiple future tokens as an auxiliary objective was formalized by \citet{gloeckle2024metamtp}, who demonstrated that training LLMs to simultaneously predict $k$ future tokens improves sample efficiency and downstream task performance. This idea has since been adopted at scale: DeepSeek-V3\&V4~\cite{liu2024deepseekv3,deepseekai2026deepseekv4}, LongCat-Flash~\cite{meituan2025longcatflashtechnicalreport} and StepFun-3.5-Flash~\cite{huang2026step35flash} incorporate MTP with cross-entropy supervision. MiMo~\cite{coreteam2025mimo} and Qwen3-Next~\cite{qwen3} employ similar multi-token heads. On the efficiency side, FastMTP~\cite{cai2025fastmtp} aligns MTP training with its recursive inference pattern, achieving significant speedups through self-distillation and dynamic vocabulary compression. The MTP heads can serve as draft models for speculative decoding~\cite{leviathan2023fast,chen2023accelerating} at inference time. These stronger representations plus faster inference make MTP an increasingly standard component in modern LLM architectures.

\paragraph{Reinforcement Learning Post-Training.}
Reinforcement Learning from Verifiable Rewards (RLVR) has become the standard paradigm for enhancing the reasoning capabilities of LLMs. Proximal Policy Optimization (PPO)~\cite{schulman2017ppo} established the foundation, using clipped surrogate objectives with a value function baseline and KL regularization against a reference policy. Group relative policy optimization (GRPO)~\cite{shao2024deepseekmath} simplifies this by eliminating the critic network and computing advantages via group-level reward normalization. Decoupled clip and dynamic sampling policy optimization (DAPO)~\cite{yu2025dapo} further refines the paradigm with clip-higher, dynamic sampling, and overlong reward shaping to improve training stability on mathematical reasoning tasks. Group sequence policy optimization (GSPO)~\cite{zheng2025gspo} reduces variance during large-scale training by shifting the optimization objective from tokens to sequences. Despite these innovations, how to jointly train RL and MTP remains largely unexplored.

\paragraph{Joint MTP-RL Training.}
Although MTP is a widely used module and optimized during pretraining, its role during RL post-training remains largely unresolved. The RL training frameworks, such as veRL~\cite{sheng2025verl} and slime~\cite{slime}, document that backpropagates MTP gradients to the main model causes severe degradation and recommend gradient detachment as the stable default~\cite{verl_mtp_docs,zhao2024slimetutorial}. Among released models, GLM-5~\cite{glm5team2026glm5} explicitly detaches MTP during RL. Nemotron-3 Super~\cite{nvidia_nemotron_3_super} adopts a healing approach where MTP is fine-tuned after RL with the main model frozen. And Composer-2~\cite{research2026composer2} similarly employs detached cross-entropy training for MTP. Several other models that use MTP (e.g., DeepSeek-V3\&V4~\cite{liu2024deepseekv3,deepseekai2026deepseekv4}, Qwen3-Next~\cite{qwen3}, StepFun-3.5-Flash~\cite{huang2026step35flash}) have not reported their RL-stage MTP strategy. To our knowledge, no prior work has provided a theoretical explanation for why joint MTP-RL training fails or proposed a principled method to enable their joint training.

\section{Joint MTP-RL Training}
\label{sec:methodology}

\begin{figure*}[t]
\centering
\begin{subfigure}[b]{0.32\linewidth}
\includegraphics[width=\linewidth]{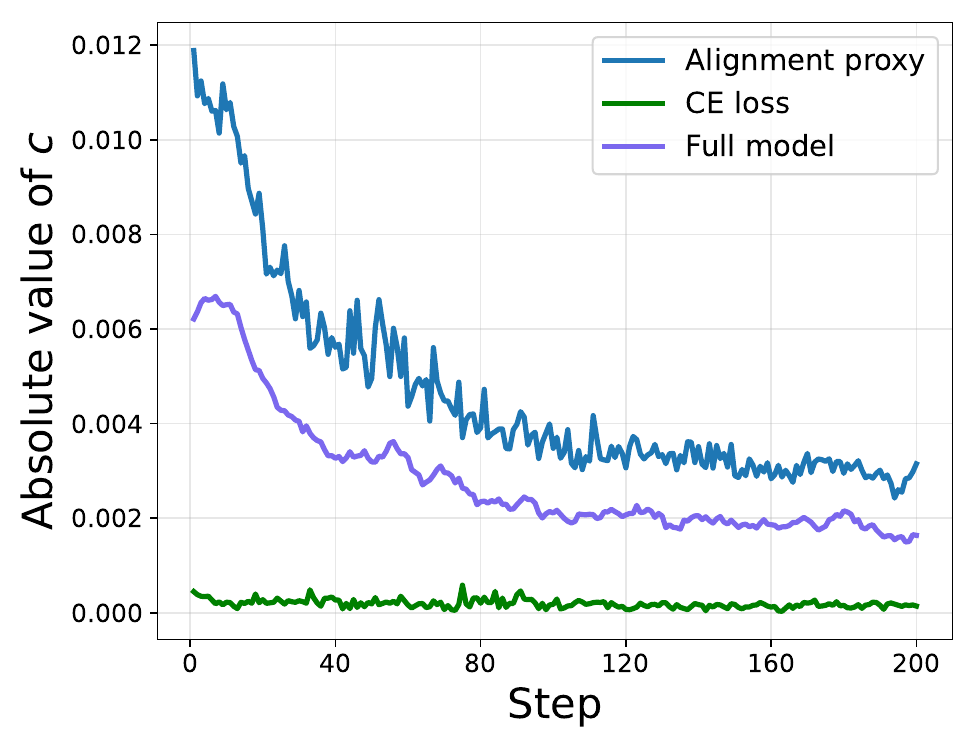}
\caption{}
\end{subfigure}\hfill
\begin{subfigure}[b]{0.32\linewidth}
\includegraphics[width=\linewidth]{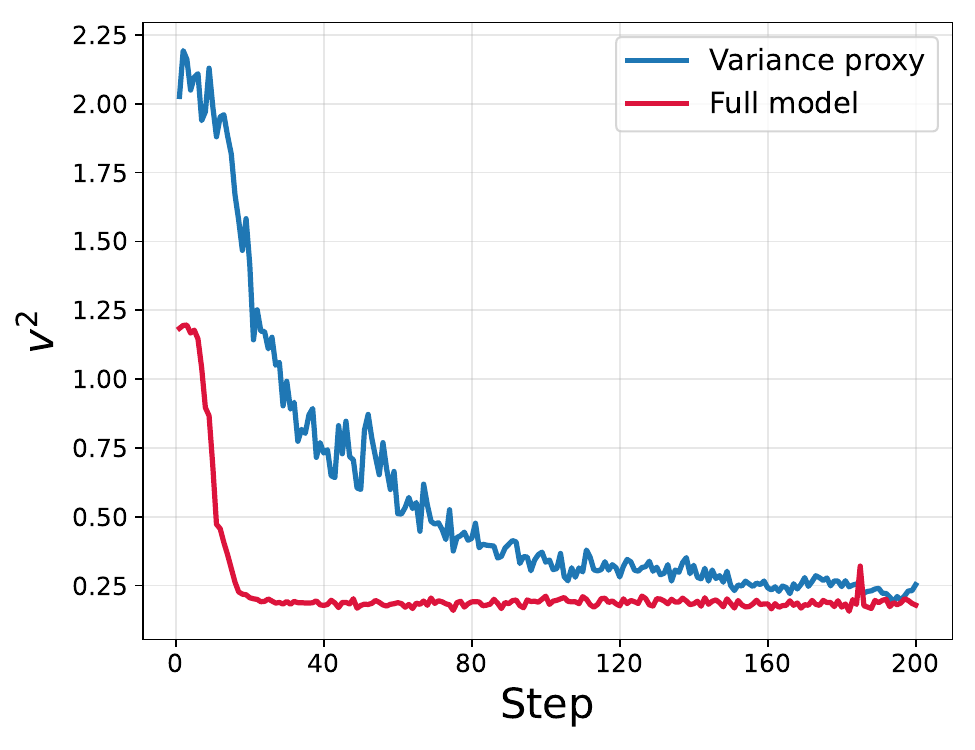}
\caption{}
\end{subfigure}\hfill
\begin{subfigure}[b]{0.32\linewidth}
\includegraphics[width=\linewidth]{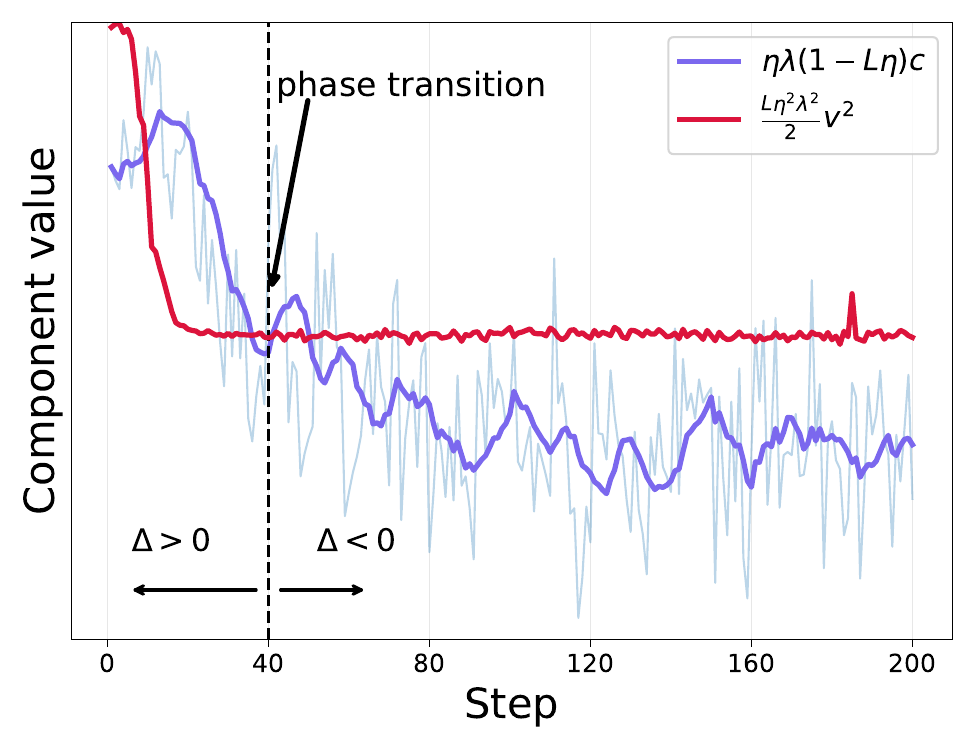}
\caption{}
\end{subfigure}
\vspace{-8pt}
\caption{Illustration of the training dynamics of coefficient components. (a) Curves of $|c|$ of full model, CE loss, and alignment proxy. While the correlation of CE loss (green curve) remains near zero, the full model correlation of policy loss declines steadily. The proxy $|\hat{c}_t|$ effectively captures the drift trends. (b) Curves of $v^2$ of full model and variance proxy. $\hat{v}_t^2$ also captures $v^2$ and stabilizes around a similar magnitude in the later stage. (c) Curves of first-order correlation term and second-order penalty term. The first-order correlation term decays while the second-order penalty stays persistent. Their crossing flips $\Delta_{\mathrm{MTP}}$ from positive gain to negative gain.}
\label{fig:main_analysis}
\end{figure*}

In this section, we first establish a joint training optimization framework to analyze joint  MTP-RL training (\S\ref{sec:framework}). We then apply this framework to explain the behavior of three MTP regimes (\S\ref{sec:three_regimes}). Next, we identify the degradation mechanism behind the policy loss regime (\S\ref{sec:late_stage}). Finally, we derive Optimal Coefficient Calibration (OCC), an adaptive procedure that calibrates MTP coefficient throughout training (\S\ref{sec:adaptive}).

\subsection{Theoretical Framework: Effect of MTP on RL Objective}
\label{sec:framework}

Consider the RL objective of maximizing the expected reward $J(\theta)$. For an $L$-smooth objective function~\cite{schulman2015trpo,zhang2020global}, there exists a constant $L > 0$ such that for all $\theta, \theta'$:
\begin{equation}
\label{eq:smooth}
J(\theta') \geq J(\theta) + \langle \nabla J(\theta), \theta' - \theta \rangle - \frac{L}{2} \| \theta' - \theta \|^2,
\end{equation}
where $\langle\cdot, \cdot\rangle$ stands for dot product. When MTP is introduced, the parameter update yields:
\begin{equation}
\label{eq:update_rule}
\theta_{t+1} = \theta_t + \eta \left( \nabla J(\theta_t) + \lambda \nabla J_{\mathrm{MTP}}(\theta_t) \right),
\end{equation}
where $\eta$ is the learning rate and $\lambda$ is the MTP loss coefficient. For notational convenience, we write $g_{\mathrm{RL}} \triangleq \nabla J(\theta_t)$ and $g_{\mathrm{MTP}} \triangleq \nabla J_{\mathrm{MTP}}(\theta_t)$.

Substituting Eq.~\eqref{eq:update_rule} into inequality~\eqref{eq:smooth}, we obtain the per-step improvement lower bound:
\begin{equation}
\label{eq:lower_bound}
J(\theta_{t+1}) \geq J(\theta_t) + \underbrace{\eta\left(1 - \frac{L\eta}{2}\right) \|g_{\mathrm{RL}}\|^2}_{\text{standard RL improvement}} + \Delta_{\mathrm{MTP}},
\end{equation}
where $\Delta_{\mathrm{MTP}}$ captures the effect of MTP:
\begin{align}
\label{eq:delta_mtp}
\Delta_{\mathrm{MTP}} =& \;\underbrace{\eta\lambda(1 - L\eta)\langle g_{\mathrm{RL}}, g_{\mathrm{MTP}} \rangle}_{\text{first-order correlation term}} \nonumber\\
&- \underbrace{\frac{L\eta^2\lambda^2}{2}\|g_{\mathrm{MTP}}\|^2}_{\text{second-order penalty term}}.
\end{align}

This decomposition shows that the effect of MTP on $J(\theta)$ is governed by two terms: the first term reflects the directional correlation between RL and MTP gradients, while the second term (always non-positive) represents the per-step perturbation introduced by the MTP gradient. Whether MTP improves or degrades $J(\theta)$ is determined by the relative magnitude of these two terms.

\subsection{Analysis of Three Training Regimes}
\label{sec:three_regimes}

We now apply Eq.~\eqref{eq:delta_mtp} to explain the behavior of three MTP training regimes.

\paragraph{Regime 1: Detach.}
When MTP gradients (usually from cross-entropy) are detached, $g_{\mathrm{MTP}} = \mathbf{0}$ w.r.t the main model, yielding $\Delta_{\mathrm{MTP}} = 0$. MTP does not affect the main model.

\paragraph{Regime 2: Cross-Entropy Loss.}
When MTP cross-entropy loss is backpropagated into the main model, its gradient yields $g_{\mathrm{CE}} = -\nabla \mathcal{L}_{\mathrm{CE}} = \frac{1}{B}\sum_{i}\nabla_\theta \log\pi(y_i \mid x_i)$\footnote{The sign of $\nabla\mathcal{L}$ is "$-$" because the RL gradient is an \textit{ascent} direction on $J$ while CE is \textit{descent} on $\mathcal{L}$.}, while the RL gradient is $g_{\mathrm{RL}} = \frac{1}{B}\sum_{i} A_i \nabla_\theta \log\pi(y_i \mid x_i)$, where $A_i$ denotes the advantage estimate for sample $i$.

A key property of $A_i$ is \emph{zero-mean}: $\frac{1}{B}\sum_i A_i \approx 0$~\cite{shao2024deepseekmath}. Under the assumption that $A_i$ are approximately independent of the corresponding per-sample gradient norms $\|\nabla\log\pi_i\|^2$, the expected first-order correlation term satisfies:
\begin{equation}
\label{eq:ce_first_order}
\mathbb{E}\left[\langle g_{\mathrm{RL}}, g_{\mathrm{CE}} \rangle\right] \approx \frac{1}{B^2}\sum_{i} A_i \|\nabla\log\pi_i\|^2 \approx 0.
\end{equation}
Since the first-order term is zero in expectation, the second-order penalty dominates:
\begin{equation}
\label{eq:ce_delta}
\Delta_{\mathrm{MTP}}^{\mathrm{CE}} \approx -\frac{L\eta^2\lambda^2}{2}\|\nabla\mathcal{L}_{\mathrm{CE}}\|^2 < 0.
\end{equation}

As shown by the green curve in Figure~\ref{fig:main_analysis}~(a), the correlation of CE loss remains near zero. While in Figure~\ref{fig:main_analysis}~(b), the variance remains substantial. This persistent variance introduces perturbation into $J(\theta)$ and inducing non-negligible performance degradation. This can be intuitively understood by noting that, RL amplifies high-reward samples and suppresses low-reward ones, whereas cross-entropy treats all samples equally regardless of reward values. Consequently, the two objectives conflict in directions and finally hurt training.

\paragraph{Regime 3: Policy Loss.}
We argue that MTP should adopt the same objective formulation as the main model (e.g., advantage, trust region, and credit assignment), as $g_{\mathrm{MTP}} = \nabla J_{\mathrm{MTP}}$ points in a direction roughly consistent with the main RL objective. Denoting the gradient inner product as $c \triangleq \langle g_{\mathrm{RL}}, g_{\mathrm{MTP}} \rangle$ and the squared norm of the MTP gradient as $v^2 \triangleq \|g_{\mathrm{MTP}}\|^2$, Eq.~\eqref{eq:delta_mtp} yields:
\begin{equation}
\label{eq:pa_delta}
\Delta_{\mathrm{MTP}}^{\mathrm{PL}} = \eta\lambda(1 - L\eta)\, c - \frac{L\eta^2\lambda^2}{2}\, v^2.
\end{equation}
This is positive when the correlation is larger than the penalty:
\begin{equation}
\label{eq:pa_condition}
\begin{aligned}
&\eta\lambda(1 - L\eta)\, c > \frac{L\eta^2\lambda^2}{2}\, v^2 \\
&\qquad\Rightarrow\quad \lambda < \frac{2(1-L\eta)\,c}{L\eta\, v^2}.
\end{aligned}
\end{equation}

Under this condition, policy loss improves the RL policy improvement lower bound.

\subsection{Degradation Behind Policy Loss}
\label{sec:late_stage}

\begin{figure}[t]
\centering
\includegraphics[width=\linewidth]{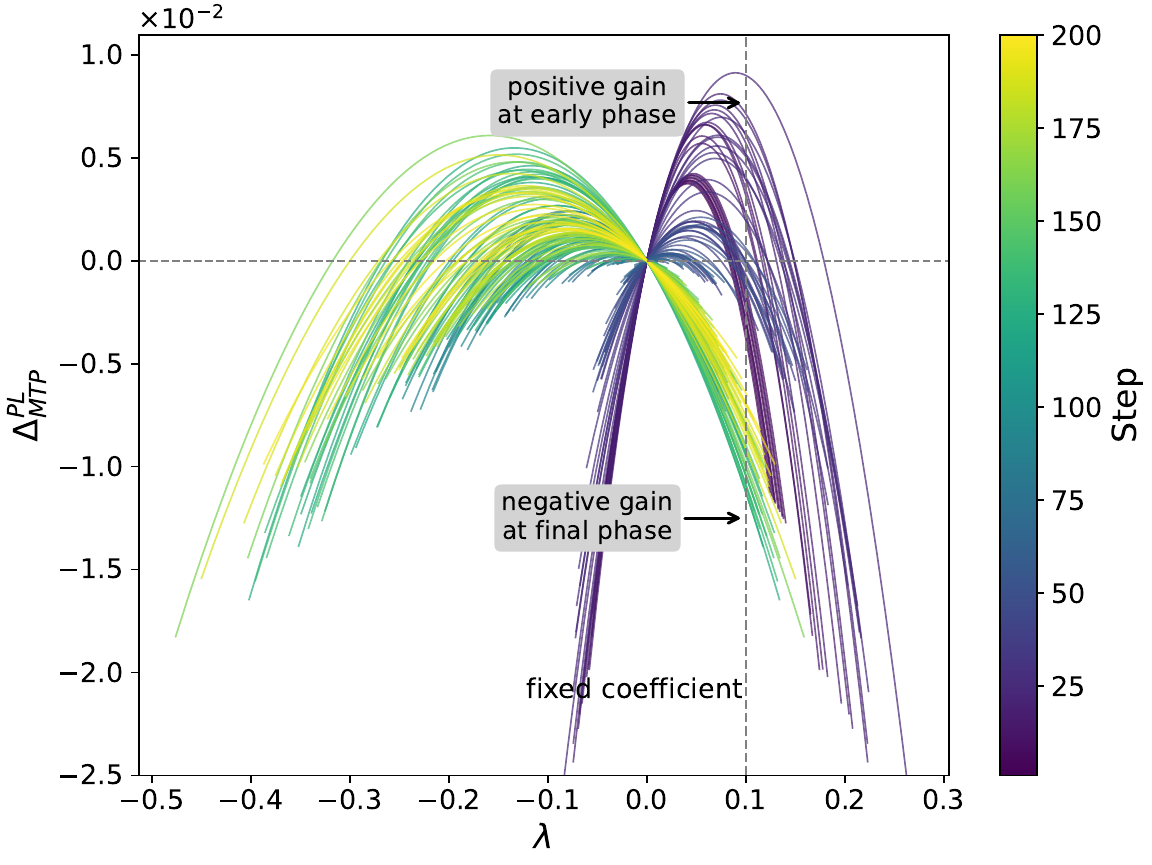}
\caption{Evolution of the policy-aligned gain across training steps (Eq.~\eqref{eq:quadratic}). Each curve is a downward-opening parabola in $\lambda$, coloured from early (dark purple) to late (bright yellow) steps. The vertical dashed line marks a fixed $\lambda$: at early steps, the intersection point of $\lambda$ and the parabola lies above zero (\emph{positive gain}), but as training progresses the parabola drifts so that the intersection point falls under zero (\emph{negative gain}).}
\label{fig:parabola_plot}
\end{figure}

Although policy loss is theoretically capable of improving $J(\theta)$, empirical observation in Figure~\ref{fig:aime24acc} reveals that it initially outperforms the Detach baseline (48.9 vs.~48.3 at 30 step), but later drops below it (38.9 vs.~45.3 at last step). We explain this behavior by analyzing how the two terms in Eq.~\eqref{eq:pa_delta} drift over the course of training.

\paragraph{Early phase: correlation dominance.} As the model is far from convergence, gradients are obtained in a steep region of the loss landscape~\cite{li2018visualizing}. Both the RL and MTP objectives produce correlated gradients. As shown by the purple curves in Figure~\ref{fig:main_analysis}~(a) and (c), $c$ is large and positive, so the correlation term dominates. Consequently, $\Delta_{\mathrm{MTP}} > 0$ and MTP improves RL training\footnote{We note that Figure~\ref{fig:main_analysis}~(a) plots $|c|$ rather than $c$. The raw $c$ is a signed quantity that can be positive or negative. Averaging the signed values produces a curve that hovers near zero. Taking the absolute value preserves the magnitude, making the curve visible.}.

\paragraph{Late phase: penalty dominance.} As the model enters a flat region, the RL and MTP gradients diverge in direction, leading to reduced correlation, while the MTP continues to produce substantial gradient norm~\cite{zhang2026beyond}, so $v^2$ remains large. As shown in Figure~\ref{fig:main_analysis}~(c), this asymmetry triggers a phase transition:
\begin{equation}
\label{eq:phase_transition}
\underbrace{\eta\lambda(1-L\eta)\,c}_{\text{weakens}} < \underbrace{\frac{L\eta^2\lambda^2}{2}\, v^2}_{\text{persists}},
\end{equation}
causing $\Delta_{\mathrm{MTP}}$ to flip from positive to negative, and MTP begins to degrade $J(\theta)$.

Figure~\ref{fig:main_analysis}~(c) shows this phase transition. During early training, correlation is strongly positive, indicating that RL and MTP gradients point in similar directions. As training progresses, correlation decays, causing the degradation of $J(\theta)$. 

\subsection{Optimal Coefficient Calibration (OCC)}
\label{sec:adaptive}

To enable an adaptive coefficient calibration that tracks the drift during training, we treat $\Delta_{\mathrm{MTP}}$ as a downward-opening parabola of $\lambda$:
\begin{equation}
\label{eq:quadratic}
\Delta_{\mathrm{MTP}}(\lambda) = \eta(1 - L\eta)\, c \cdot \lambda - \frac{L\eta^2}{2}\, v^2 \cdot \lambda^2,
\end{equation}

Figure~\ref{fig:parabola_plot} illustrates the evolutionary trajectory of Eq.~\eqref{eq:quadratic} across training steps. Under a fixed coefficient, the gain exhibits a drift: during the early phase, the gain remains positive. However, as the training progresses, the gain drifts from positive to negative. This analysis yields a key insight: the MTP coefficient should not be fixed but should adaptively calibrate the drift.

\paragraph{Closed-form optimal coefficient.}
At every step, we set $\lambda$ to the optimal point of current parabola, thereby maximizing the per-step improvement. The maximum of Eq.~\eqref{eq:quadratic} is attained at:
\begin{equation}
\label{eq:lambda_star}
\lambda^* = \frac{(1-L\eta)\,c}{L\eta\, v^2} = \frac{1-L\eta}{L\eta} \cdot \frac{c}{v^2}.
\end{equation}

Since the smoothness constant $L$ is unknown in practice, the prefactor $\frac{1-L\eta}{L\eta}$ cannot be computed directly. However, since the constant is fixed during training, it can be absorbed into a global scaling factor. The theoretically meaningful variable that governs the optimal coefficient is the ratio $c / v^2$.

\paragraph{Log-probability gradient proxy.}
Computing full-model gradients is prohibitively expensive in large-scale distributed training systems~\cite{megatron-lm}. Inspired by OTB~\cite{li2026otb} and FIPO~\cite{ma2026fipo}, we use the change in log-probability under small-step updates as a proxy for the gradient:
\begin{equation}
\label{eq:proxy}
\langle g_1, g_2 \rangle \propto \langle \delta_1, \delta_2 \rangle, \qquad \|g\|^2 \propto \|\delta\|^2,
\end{equation}
where $\delta \triangleq \log\pi_\theta - \log\pi_{\mathrm{old}}$ denotes the log-probability change between current and old policy within small-step updates. This proxy calculation is negligible compared to computing the whole gradient but it serves as a good indicator, as shown by the blue curves in Figure~\ref{fig:main_analysis}. We define two online statistics computed over each training batch:

\noindent\textbf{(1) Alignment proxy:} $\hat{c}_t = \langle \delta_{\mathrm{RL}}, \delta_{\mathrm{MTP}} \rangle$, estimating the gradient inner product $c$.

\noindent\textbf{(2) Variance proxy:} $\hat{v}_t^2 = \|\delta_{\mathrm{MTP}}\|^2$, estimating the auxiliary gradient norm $v^2$.

The optimal coefficient per-step is given by:
\begin{equation}
\label{eq:adaptive_weight}
\lambda_t = \lambda_{+}\cdot\frac{\hat{c}_t}{\hat{v}_t^2 + \epsilon},
\end{equation}
where $\epsilon > 0$ is a small constant to avoid division by zero and $\lambda_{+}$ is a predefined ratio that absorbs the unknown smoothness prefactor.

\begin{table*}[t]
\centering
\caption{Main results on mathematical reasoning benchmarks across models and algorithms. We report avg@32 accuracy (\%). Best results are in \textbf{bold}; second best are \underline{underlined}.}
\footnotesize
\label{tab:main_results}
\renewcommand\arraystretch{1.12}
\resizebox{0.88\linewidth}{!}{
\begin{tabular}{lcccccc|c}
\toprule
\textbf{Regime} & \textbf{AIME24} & \textbf{AIME25} & \textbf{AMC} & \textbf{MATH} & \textbf{Minerva} & \textbf{Olympiad} & \textbf{Avg.} \\
\midrule
\rowcolor{gray!20} \multicolumn{8}{c}{\emph{\scalebox{0.8}{MiMo-7B-RL w/ DAPO}}} \\
Detach & \underline{45.3} & \underline{36.7} & \underline{81.9} & \underline{90.0} & \underline{39.7} & \underline{60.1} & \underline{58.9} \\
CE Loss & 16.1 & 33.3 & 67.5 & 83.2 & 33.8 & 52.3 & 47.7 \\
Policy Loss & 38.9 & \underline{36.7} & 79.5 & 89.8 & \underline{39.7} & 59.7 & 57.4 \\
OCC (Ours) & \textbf{45.9} & \textbf{46.7} & \textbf{83.1} & \textbf{91.0} & \textbf{43.4} & \textbf{60.3} & \textbf{61.7} \\
\midrule
\rowcolor{gray!20} \multicolumn{8}{c}{\emph{\scalebox{0.8}{MiMo-7B-RL w/ GSPO}}} \\
Detach     & \underline{43.4} & 33.3 & \underline{79.5} & \underline{90.6} & \underline{40.1} & \underline{59.7} & \underline{57.8} \\
CE Loss     & 10.3 & 16.7 & 63.9 & 81.8 & 29.0 & 44.1 & 50.0 \\
Policy Loss  & 37.4 & \underline{34.9} & 78.6 & 88.9 & 38.4 & 58.7 & 56.2 \\
OCC (Ours) & \textbf{47.6} & \textbf{36.2} & \textbf{84.3} & \textbf{91.8} & \textbf{40.4} & \textbf{60.1} & \textbf{60.1} \\
\midrule
\rowcolor{gray!20} \multicolumn{8}{c}{\emph{\scalebox{0.8}{GLM-4.5-Air w/ DAPO}}} \\
Detach     & \underline{63.8} & \underline{50.0} & \underline{84.3} & 92.8 & 40.4 & \underline{63.9} & \underline{65.9} \\
CE Loss     & 50.2 & 33.3 & 68.7 & 87.8 & 40.2 & 48.6 & 54.8 \\
Policy Loss  & 57.3 & 49.9 & 83.2 & \underline{93.1} & \underline{40.5} & 62.2 & 64.4 \\
OCC (Ours) & \textbf{64.7} & \textbf{52.8} & \textbf{85.4} & \textbf{93.3} & \textbf{42.7} & \textbf{66.6} & \textbf{67.6} \\
\bottomrule
\end{tabular}
}
\end{table*}
\section{Experiments}
\label{sec:experiments}

\begin{figure*}[t]
\centering
\begin{minipage}[t]{0.24\linewidth}
  \centering
  \includegraphics[width=\linewidth]{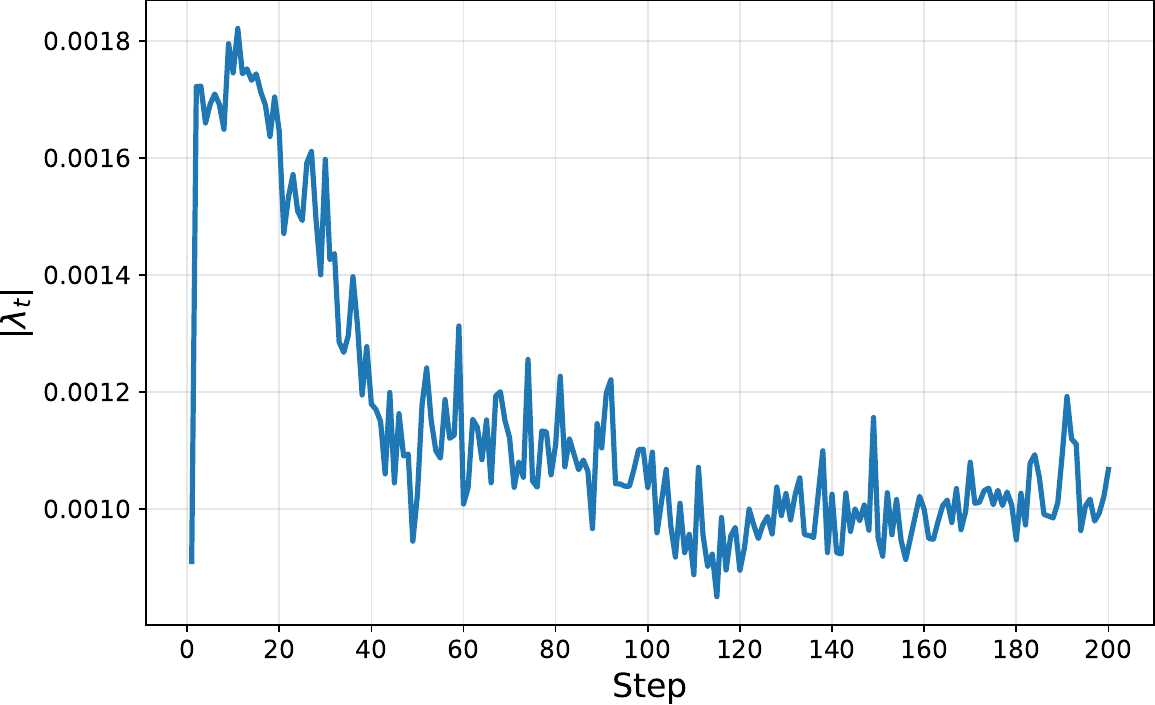}
  \captionof{figure}{Curve of the absolute value of OCC coefficient $\lambda_t$.}
  \label{fig:abs_mtp_scaling}
\end{minipage}\hfill
\begin{minipage}[t]{0.24\linewidth}
  \centering
  \includegraphics[width=\linewidth]{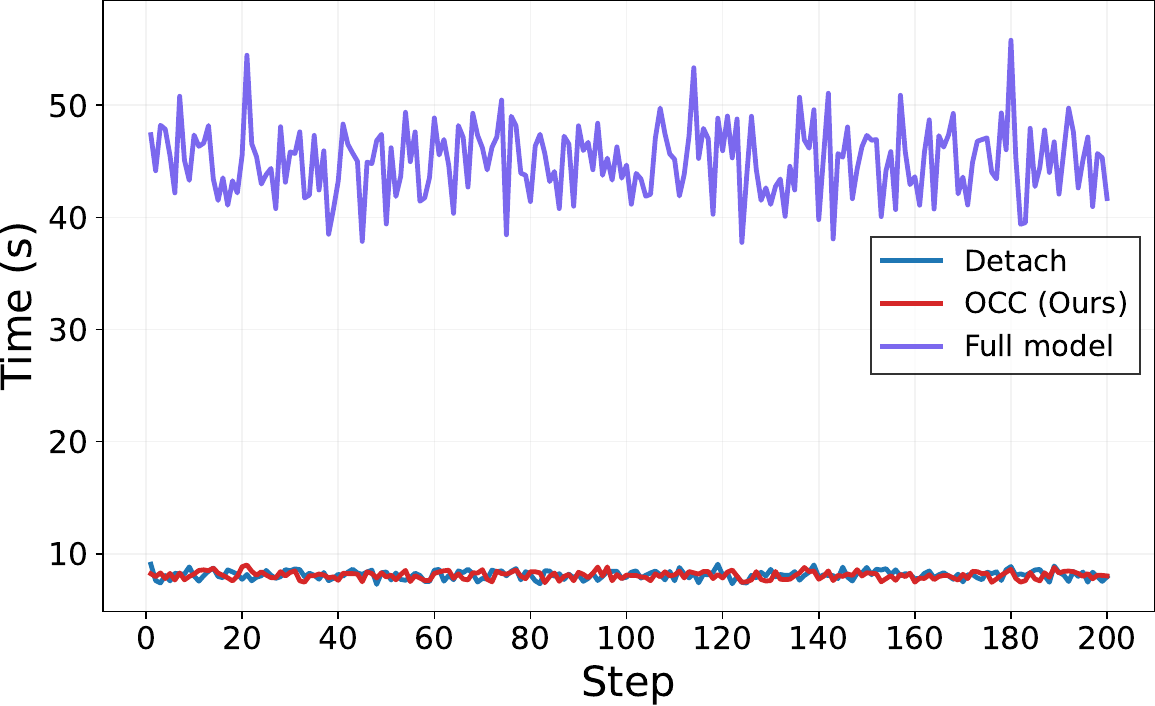}
  \captionof{figure}{Wall-clock training time of OCC, Detach and Full-model.}
  \label{fig:time_update_parameter}
\end{minipage}\hfill
\begin{minipage}[t]{0.24\linewidth}
  \centering
  \includegraphics[width=\linewidth]{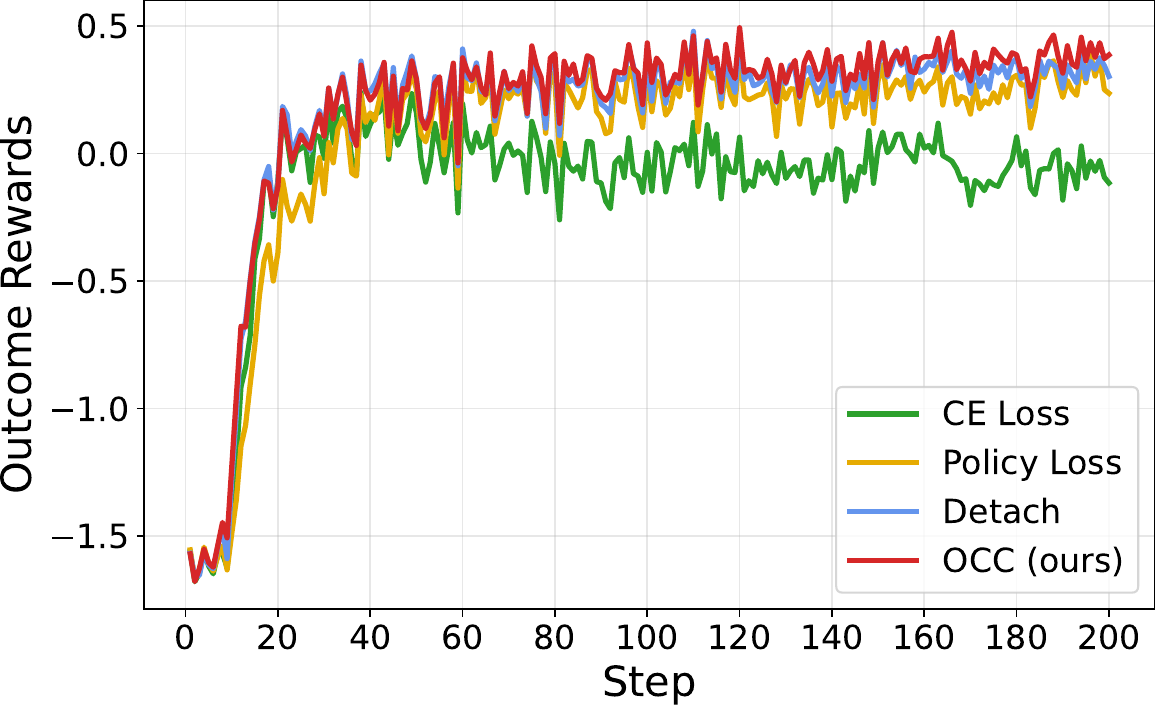}
  \captionof{figure}{Training reward of CE Loss, Detach, Policy Loss and OCC.}
  \label{fig:reward_comparison}
\end{minipage}\hfill
\begin{minipage}[t]{0.24\linewidth}
  \centering
  \includegraphics[width=\linewidth]{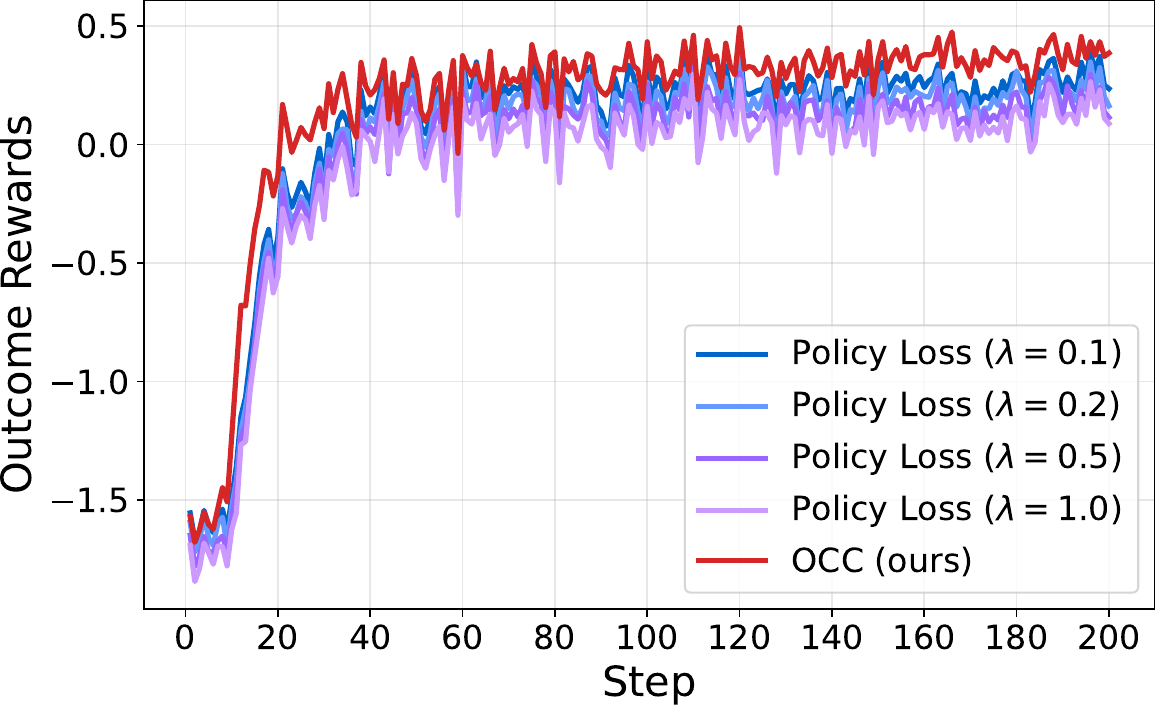}
  \captionof{figure}{Training reward of the Policy Loss with different coefficient.}
  \label{fig:reward_ablation}
\end{minipage}
\end{figure*}

\subsection{Experimental Setup}
\label{sec:exp_setup}

\paragraph{Training data.}
Building upon the experimental frameworks established in previous attempts~\cite{verl_mtp_docs, zhao2024slimetutorial}, we utilize DAPO-Math-17k~\cite{yu2025dapo} as training dataset. This specialized corpus consists of 17k mathematical reasoning prompts, designed to challenge and evaluate the logical depth of LLMs.

\paragraph{RL algorithms.}
We use DAPO~\cite{yu2025dapo} as the primary RL algorithm, aligning with the previous attempts from veRL~\cite{sheng2025verl} and slime~\cite{slime}. To demonstrate that OCC is algorithm-agnostic, we additionally report results with GSPO~\cite{zheng2025gspo}.

\paragraph{Evaluation benchmarks.}
We evaluate on multiple widely-adopted mathematical reasoning benchmarks: AIME24~\cite{aime24}, AIME25~\cite{aime25}, AMC~\cite{li2024amc}, MATH~\cite{hendrycks2021math}, Minerva~\cite{lewkowycz2022minerva} and OlympiadBench~\cite{he2024olympiadbench}. Answer correctness is verified using the DeepScaler~\cite{deepscaler2025} evaluation framework. Results are reported as the average accuracy over 32 runs.

\paragraph{Baseline methods.}
We use MiMo-7B-RL~\cite{coreteam2025mimo} with MTP head as the base model. Also, we additionally use GLM-4.5-Air~\cite{glm5team2026glm5}, a 106B-A12B Mixture-of-Experts model with MTP head to verify generalization across model scales and architectures. We compare our method with the following regimes from the theoretical framework: (1) \textbf{Detach}: MTP heads participate in forward computation but their gradients are detached from the main model. (2) \textbf{CE Loss}: MTP backpropagates cross-entropy gradients into the main model. (3) \textbf{Policy Loss}: MTP uses the same RL objective as the main model with a fixed coefficient $\lambda$. (4) \textbf{OCC (Ours)}: dynamically calibrates optimal coefficient according to Eq.~\eqref{eq:adaptive_weight} with log-probability proxy.

\paragraph{Implementation details.}
All experiments are conducted using veRL~\cite{sheng2025verl} framework. We train for 200 steps with batch size 128 and learning rate 1e-6. The rollout generation uses a temperature of 1.0. For CE loss, we set $\lambda = 0.1$ following the commonly used auxiliary loss coefficient in prior work. For policy loss, we similarly use $\lambda = 0.1$. For OCC, we set $\lambda_{+}=1.0$ and $\epsilon=10^{-8}$ as the defaults. All other hyperparameters follow the default configuration in veRL. Training is performed on 128 NVIDIA H20 GPUs.

\subsection{Main Results}
\label{sec:main_results}

Table~\ref{tab:main_results} presents the results on the benchmarks using MiMo-7B-RL with both DAPO and GSPO. We highlight three findings:

\paragraph{CE loss consistently underperforms Detach.}
As shown in Table~\ref{tab:main_results} and Figure~\ref{fig:aime24acc}, across all benchmarks, CE loss falls far below Detach (47.7 vs.~58.9 on average), with the largest gap on AIME24 (-29.2 points). This empirically confirms the theoretical analysis of \S\ref{sec:three_regimes}: the first-order correlation term vanishes, leaving only the negative second-order penalty. The cross-entropy objective thus injects uncorrelated perturbation into the RL update and systematically suppresses the policy improvement lower bound.

\paragraph{OCC consistently matches or exceeds Detach.}
OCC achieves the best average accuracy (61.7), exceeding Detach by +2.8 points and policy loss by +4.3 points. The largest margin appears on AIME25 (+10.0 points over detach and policy loss), indicating that OCC captures the drift effectively and adjust the coefficient to improve the training. Consistent gains appear MATH (+1.0), AMC (+1.2), and Minerva (+3.7). The same gain is visible in the training-reward trajectories in Figure~\ref{fig:reward_comparison}: CE loss degrades below the other regimes. Policy loss peaks then declines, consistent with the rise-then-fall narrative in \S\ref{sec:three_regimes}. OCC attains the highest reward, showing the capability to improve joint MTP-RL training.

\paragraph{Generalization across RL algorithms.}
Using GSPO~\cite{zheng2025gspo} with MiMo-7B-RL, the same result is shown in the middle of Table~\ref{tab:main_results}: CE loss remains worst, policy loss underperforms Detach, and OCC is uniformly best, reaching 60.1 average accuracy versus 57.8 for Detach (+2.3). As with DAPO, consistent gains appear across benchmarks, indicating that OCC's benefit in joint MTP-RL training.

\paragraph{Generalization across base models.}

Table~\ref{tab:main_results} also evaluates the regimes on GLM-4.5-Air~\cite{glm5team2026glm5}, a 106B-A12B MoE model. The results hold: OCC reaches 67.6 average (vs.~65.9 for Detach, +1.7), with performance gains on every benchmark. This confirms that OCC generalizes across model scales, remaining effective from a dense 7B model up to a 106B MoE.

\subsection{Ablation Study}
\label{sec:ablation}

\paragraph{Fidelity of log-probability gradient proxy.}
The log-probability gradient proxies $\hat{c}_t$ and $\hat{v}_t^2$ can reflect the behaviour of the full model gradient, as fully discussed in previous works~\cite{li2026otb,ma2026fipo}. Figure~\ref{fig:main_analysis} provides empirical support: the proxy $\hat{c}_t$ in panel~(a) exhibits the consistent decay predicted by the analysis for the gradient correlation term, transitioning from a positive correlation-dominant phase to near-zero in the late stage; the variance proxy $\hat{v}_t^2$ in panel~(b) remains bounded away from zero and stabilizes around $0.25$, mirroring the persistent MTP-gradient variance that drives the second-order penalty. 

\begin{figure}[h]
\centering
\includegraphics[width=\linewidth]{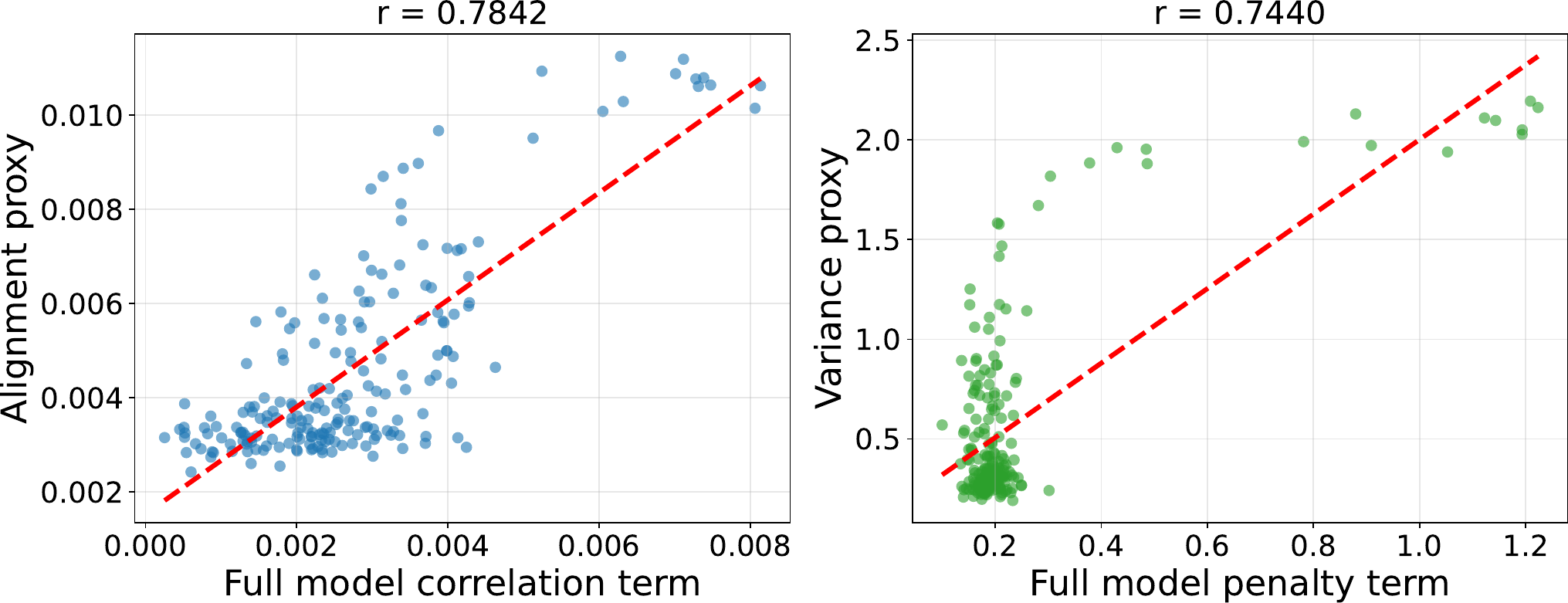}
\caption{Relationship betwen the log-probability proxies and the true full-model gradient quantities. Each point corresponds to one training step. \textbf{Left:} the alignment proxy plotted against the full-model first-order correlation. \textbf{Right:} the variance proxy plotted against the full-model second-order penalty. The proxies exhibit highly positive correlation with the true gradient quantities ($r$ is annotated above).}
\label{fig:correlation_scatter}
\end{figure}

Figure~\ref{fig:correlation_scatter} further validates the fidelity of these proxies by directly comparing them against the full-model gradient quantities: the alignment proxy $\hat{c}_t$ shows strong positive correlation with the true $c$ (left), and the variance proxy $\hat{v}_t^2$ similarly correlates with the full-model second-order penalty $v^2$ (right). In both panels the linear regression fit confirms that the cheap log-probability proxy is a faithful substitute for the expensive full gradient statistics that drive OCC.

\begin{table*}[t]
\centering
\footnotesize
\caption{Comparison of policy loss with $\lambda \in \{0.1, 0.2, 0.5, 1.0\}$ against Detach and OCC.}
\label{tab:pa_static_sweep}
\resizebox{0.88\linewidth}{!}{
\begin{tabular}{lcccccc|c}
\toprule
\textbf{Method} & \textbf{AIME24} & \textbf{AIME25} & \textbf{AMC} & \textbf{MATH} & \textbf{Minerva} & \textbf{Olympiad} & \textbf{Avg.} \\
\midrule
Detach                    & 45.3 & 36.7 & 81.9 & 90.0 & 39.7 & 60.1 & 58.9 \\
\midrule
Policy Loss ($\lambda{=}0.1$)& 38.9 & 36.7 & 79.5 & 89.8 & 39.7 & 59.7 & 57.4 \\
Policy Loss ($\lambda{=}0.2$) & 38.7 & 36.6 & 80.7 & 89.9 & 39.5 & 59.8 & 57.5 \\
Policy Loss ($\lambda{=}0.5$) & 38.9 & 36.7 & 79.5 & 89.8 & 39.7 & 59.7 & 57.4 \\
Policy Loss ($\lambda{=}1.0$) & 38.8 & 36.4 & 78.1 & 88.3 & 40.6 & 57.1 & 56.6 \\
\midrule
OCC (Ours)                & \textbf{45.9} & \textbf{46.7} & \textbf{83.1} & \textbf{91.0} & \textbf{43.4} & \textbf{60.3} & \textbf{61.7} \\
\bottomrule
\end{tabular}
}
\end{table*}

\paragraph{OCC automatically adjust $\lambda_t$.}

Figure~\ref{fig:abs_mtp_scaling} traces the adaptive coefficient $\lambda_t = \lambda_{+}\cdot\hat{c}_t/(\hat{v}_t^2+\epsilon)$ produced by OCC. The trajectory follows the mechanism modeled by Eq.~\eqref{eq:adaptive_weight} and shown by Figure~\ref{fig:main_analysis}: $\lambda_t$ is large during the correlation-dominant phase, then decays to near zero once $\hat{c}_t$ drifts. The curve shows that OCC captures the phase transition during training, end effectively adjust $\lambda_t$ to ensure the positive gain during training.

\paragraph{Training Time Overhead}
We compare the per-step wall-clock training time under Detach, OCC, and full-model gradient computation. As shown in Figure~\ref{fig:time_update_parameter}, OCC achieves an average step time of $8.07$ s, comparable to Detach at $8.11$ s, whereas full-model backpropagation requires $45.06$ s, yielding a $5.6\times$ slowdown. This indicates that OCC introduces negligible overhead while avoiding the prohibitive cost of exact MTP gradient computation, thereby enabling efficient adaptive MTP weighting for large-scale distributed RL training.

\paragraph{Policy loss cannot match OCC at any fixed coefficient.}

An important question is whether the weaker performance of policy loss is due to suboptimal hyperparameter tuning. In Figure~\ref{fig:reward_comparison}, we compare the training reward of the three MTP variants with OCC. CE loss shows clear reward degradation, while Detach and policy loss both fall behind OCC in the later stage of training. OCC also achieves the highest reward than other regimes. 

Figure~\ref{fig:reward_ablation} and Table~\ref{tab:pa_static_sweep} sweep the coefficient of policy loss across $\lambda \in \{0.1, 0.2, 0.5, 1.0\}$. Results show that all static coefficient settings are surpassed by OCC on the average and on every individual benchmark, with small $\lambda$ behaving closely to Detach and large $\lambda$ collapsing in the late stage. These results show that no fixed coefficient can match OCC. The improvement from OCC therefore comes from its dynamic adjustment w.r.t the drift of gradient alignment and phase transition, rather than from a well-tuned constant coefficient.

\paragraph{Sensitivity of $\lambda_{+}$.}

The predefined ratio $\lambda_{+}$ absorbs the smoothness prefactor $\frac{1-L\eta}{L\eta}$ in Eq.~\eqref{eq:lambda_star}. Table~\ref{tab:lambda_sensitivity} presents the sensitivity of OCC to different choices of $\lambda_{+}$. While individual benchmark performance varies, selecting $\lambda_{+} = 1.0$ based on the average performance across all benchmarks yields the best overall results. This robustness arises because the adaptive ratio $\hat{c}_t / \hat{v}t^2$ already captures the dominant dynamics, and $\lambda{+}$ primarily rescales the overall magnitude.

\begin{table}[t]
\centering
\caption{Sensitivity analysis of $\lambda_{+}$. We sweep $\lambda_{+} \in \{0.1, 0.2, 0.5, 1.0, 2.0\}$ and compare the performance across various benchmarks. $\lambda_{+} = 1.0$ yields the best average performance, indicating it as the preferred choice based on overall results. Best in \textbf{bold}.}
\label{tab:lambda_sensitivity}
\setlength{\tabcolsep}{4pt}
\resizebox{\linewidth}{!}{
\begin{tabular}{lcccccc|c}
\toprule
\textbf{$\lambda_+$} & \textbf{AIME24} & \textbf{AIME25} & \textbf{AMC} & \textbf{MATH} & \textbf{Minerva} & \textbf{Olympiad} & \textbf{Avg.} \\
\midrule
0.1  & \textbf{47.0} & 43.8 & 81.7 & 90.1 & 41.5 & 59.4 & 60.6 \\
0.2 & 45.1 & 44.6 & \textbf{83.3} & 90.4 & 42.3 & 59.7 & 60.9 \\
0.5  & 45.6 & 46.2 & 82.8 & 90.7 & 42.9 & 60.0 & 61.4 \\
1.0  & 45.9 & \textbf{46.7} & 83.1 & 91.0 & \textbf{43.4} & \textbf{60.3} & \textbf{61.7} \\
2.0  & 44.8 & 45.4 & 82.0 & \textbf{91.8} & 42.1 & 59.6 & 60.9 \\
\bottomrule
\end{tabular}
}
\end{table}
\section{Conclusion}
\label{sec:conclusion}

We revisit the failure of joint MTP-RL training from an optimization perspective. Starting from an $L$-smoothness argument, we derived a per-step improvement bound that decomposes the effect of MTP into a first-order correlation term and a second-order penalty. This decomposition unifies the Detach, Cross-Entropy, and Policy Loss regimes under a single framework, and exposes a phase transition during training that no static coefficient can track. Guided by this analysis, we proposed \textbf{Optimal Coefficient Calibration (OCC)}, which sets the MTP coefficient to its closed-form optimum at every step via a log-probability proxy at negligible cost. Extensive results on multple benchmarks show that OCC consistently matches or exceeds the detach baseline, demonstrating that MTP can be safely re-introduced into RL post-training and deliver consistent gains.


\newpage

\section{Limitations}

Our theoretical analysis rests on the $L$-smoothness assumption, which may not hold exactly in non-convex deep learning landscapes; the bound provides qualitative rather than quantitative guidance. The log-probability proxy for gradient alignment is a first-order approximation that may lose fidelity under large learning rates or aggressive policy updates. Our experiments focus on mathematical reasoning with verifiable rewards; generalization to other RL post-training settings such as RLHF with learned reward models or open-ended generation remains to be validated. Finally, the predefined ratio $\lambda_{+}$ absorbs an unknown smoothness-dependent prefactor; while our ablations show robustness across a wide range, the optimal choice may vary across training configurations.

\bibliography{custom}
\appendix
\clearpage
\section{Detailed Proof of the Policy Improvement Lower Bound}
\label{app:proof}

We provide the full derivation of Eq.~\eqref{eq:lower_bound} and Eq.~\eqref{eq:delta_mtp} from the main text.

Starting from the $L$-smoothness condition:
\begin{equation}
J(\theta') \geq J(\theta) + \langle \nabla J(\theta),\; \theta' - \theta \rangle - \frac{L}{2}\|\theta' - \theta\|^2.
\end{equation}

This is the standard lower bound for $L$-smooth functions in the maximization setting, implying Lipschitz continuity of gradients. We further assume $\eta < 1/L$ so that the quadratic approximation remains valid within the update neighborhood.

Setting $\theta = \theta_t$ and $\theta' = \theta_{t+1} = \theta_t + \eta(g_{\mathrm{RL}} + \lambda g_{\mathrm{MTP}})$:
\begin{align}
J(\theta_{t+1}) \geq& \; J(\theta_t) + \eta\langle g_{\mathrm{RL}},\, g_{\mathrm{RL}} + \lambda g_{\mathrm{MTP}}\rangle \nonumber\\
&- \frac{L\eta^2}{2}\|g_{\mathrm{RL}} + \lambda g_{\mathrm{MTP}}\|^2.
\end{align}

Expanding the inner product:
\begin{equation}
\langle g_{\mathrm{RL}},\, g_{\mathrm{RL}} + \lambda g_{\mathrm{MTP}}\rangle = \|g_{\mathrm{RL}}\|^2 + \lambda\langle g_{\mathrm{RL}}, g_{\mathrm{MTP}}\rangle.
\end{equation}

Expanding the squared norm:
\begin{align}
\|g_{\mathrm{RL}} + \lambda g_{\mathrm{MTP}}\|^2 =&\; \|g_{\mathrm{RL}}\|^2 \nonumber\\
&+ 2\lambda\langle g_{\mathrm{RL}}, g_{\mathrm{MTP}}\rangle \nonumber\\
&+ \lambda^2\|g_{\mathrm{MTP}}\|^2.
\end{align}

Substituting and grouping by terms involving $\|g_{\mathrm{RL}}\|^2$, $\langle g_{\mathrm{RL}}, g_{\mathrm{MTP}}\rangle$, and $\|g_{\mathrm{MTP}}\|^2$:
\begin{align}
J(\theta_{t+1}) &\geq J(\theta_t) + \eta\|g_{\mathrm{RL}}\|^2 + \eta\lambda\langle g_{\mathrm{RL}}, g_{\mathrm{MTP}}\rangle \nonumber \\
&\quad - \frac{L\eta^2}{2}\|g_{\mathrm{RL}}\|^2 - L\eta^2\lambda\langle g_{\mathrm{RL}}, g_{\mathrm{MTP}}\rangle \nonumber \\&\quad- \frac{L\eta^2\lambda^2}{2}\|g_{\mathrm{MTP}}\|^2 \nonumber \\
&= J(\theta_t) + \eta\!\left(1 - \frac{L\eta}{2}\right)\!\|g_{\mathrm{RL}}\|^2 \nonumber \\&\quad+ \eta\lambda(1-L\eta)\langle g_{\mathrm{RL}}, g_{\mathrm{MTP}}\rangle \nonumber \\&\quad- \frac{L\eta^2\lambda^2}{2}\|g_{\mathrm{MTP}}\|^2.
\end{align}

This yields Eq.~\eqref{eq:lower_bound} with $\Delta_{\mathrm{MTP}}$ as defined in Eq.~\eqref{eq:delta_mtp}. Note that under the assumption $\eta < 1/L$, the coefficient $(1 - L\eta) > 0$ and $(1 - L\eta/2) > 1/2 > 0$, ensuring that the standard RL improvement term is strictly positive when $g_{\mathrm{RL}} \neq \mathbf{0}$.

\section{Vanishing First-Order Term under Cross-Entropy MTP}
\label{app:ce_proof}

We provide the full derivation of Eq.~\eqref{eq:ce_first_order} and Eq.~\eqref{eq:ce_delta} from the main text. Let $u_i \triangleq \nabla_\theta\log\pi(y_i\mid x_i)$ denote the per-sample log-likelihood gradient. Following the ascent convention of Eq.~\eqref{eq:update_rule}, we write
\begin{align}
g_{\mathrm{RL}}  &= \frac{1}{B}\sum_{i=1}^{B} A_i\, u_i, \nonumber\\\qquad g_{\mathrm{MTP}} &= -\nabla\mathcal{L}_{\mathrm{CE}} = \frac{1}{B}\sum_{j=1}^{B} u_j,
\end{align}

Expanding the first-order correlation term and splitting the diagonal from the cross-sample contributions:
\begin{align}
\label{eq:app_ce_expand}
\langle g_{\mathrm{RL}}, g_{\mathrm{MTP}}\rangle
 &= \frac{1}{B^2}\sum_{i,j} A_i\,\langle u_i, u_j\rangle \nonumber\\
 &= \underbrace{\frac{1}{B^2}\sum_i A_i\|u_i\|^2}_{\text{diagonal}}
   + \underbrace{\frac{1}{B^2}\sum_{i\neq j} A_i\langle u_i, u_j\rangle}_{\text{cross-sample}}.
\end{align}
In a standard RLVR batch, different prompts share little token-level overlap, so the cross-sample inner products $\langle u_i, u_j\rangle$ for $i\neq j$ are small. Summed against the zero-mean advantages, they vanish in expectation:
\begin{equation}
\mathbb{E}\!\left[\frac{1}{B^2}\sum_{i\neq j} A_i\langle u_i, u_j\rangle\right] \approx 0.
\end{equation}

Retaining only the diagonal and averaging across batches yields:
\begin{equation}
\label{eq:app_ce_diag}
\mathbb{E}\!\left[\langle g_{\mathrm{RL}}, g_{\mathrm{MTP}}\rangle\right] \;\approx\; \frac{1}{B^2}\sum_i A_i\,\|u_i\|^2.
\end{equation}

Finally, assuming $A_i$ is approximately independent of $\|u_i\|^2$ --- intuitively, $A_i$ is determined by trajectory-level reward while $\|u_i\|^2$ reflects per-token probability structure of the current policy --- the expectation factors:
\begin{equation}
\mathbb{E}\!\left[A_i\,\|u_i\|^2\right] \;=\; \mathbb{E}[A_i]\cdot\mathbb{E}\!\left[\|u_i\|^2\right] \;=\; 0,
\end{equation}
since the group-normalised advantage used in GRPO / DAPO / GSPO satisfies $\mathbb{E}[A_i]=0$ by construction.

Substituting this vanishing first-order term into Eq.~\eqref{eq:delta_mtp} recovers Eq.~\eqref{eq:ce_delta}:
\begin{equation}
\mathbb{E}\!\left[\Delta_{\mathrm{MTP}}^{\mathrm{CE}}\right] \;\approx\; -\frac{L\eta^2\lambda^2}{2}\,\mathbb{E}\!\left[\|\nabla\mathcal{L}_{\mathrm{CE}}\|^2\right] \;<\; 0,
\end{equation}
which is strictly negative whenever $\nabla\mathcal{L}_{\mathrm{CE}}\neq\mathbf{0}$. CE loss therefore injects pure perturbation into the RL update without a correlation gain, and the magnitude of the damage scales quadratically with $\lambda$. This is consistent with the flat, near-zero $|c|$ curve of CE loss in Figure~\ref{fig:main_analysis}~(a) and the large benchmark gap reported for CE loss in Table~\ref{tab:main_results}.

\section{Derivation of the Optimal Adaptive Weight}
\label{app:optimal_weight}

Treating $\Delta_{\mathrm{MTP}}$ as a function of $\lambda$:
\begin{equation}
f(\lambda) = \eta(1-L\eta)\,c\cdot\lambda - \frac{L\eta^2}{2}\,v^2\cdot\lambda^2,
\end{equation}
where $c = \langle g_{\mathrm{RL}}, g_{\mathrm{MTP}}\rangle$ and $v^2 = \|g_{\mathrm{MTP}}\|^2$. This is a concave quadratic in $\lambda$ (since $-\frac{L\eta^2}{2}v^2 < 0$). Setting $f'(\lambda) = 0$:
\begin{align}
&\eta(1-L\eta)\,c - L\eta^2 v^2 \lambda = 0 \nonumber \\ &\Longrightarrow\quad \lambda^* = \frac{(1-L\eta)\,c}{L\eta\, v^2}.
\end{align}

The maximum value of the perturbation at this optimum is:
\begin{equation}
f(\lambda^*) = \frac{\eta(1-L\eta)^2 c^2}{2L\eta\, v^2} = \frac{(1-L\eta)^2 c^2}{2L\, v^2} > 0,
\end{equation}

\section{Experiment Details}
\label{app:hyperparams}

Experiments are implemented in veRL~\cite{sheng2025verl} with the Megatron-LM training backend~\cite{megatron-lm} and SGLang rollout engine~\cite{zheng2024sglang}. The four regimes (Detach, CE loss, policy loss, OCC) share the same base configuration and differ only in (i) whether MTP gradients flow into the main model and (ii) the choice of MTP loss coefficient. GSPO runs replace the DAPO advantage estimator and surrogate loss while keeping everything else unchanged.

\paragraph{Dataset and benchmark licenses.}
We use publicly released training and evaluation resources and follow the license metadata declared by their corresponding dataset or repository pages. The DAPO-Math-17k training dataset is released under Apache-2.0. For evaluation benchmarks, AIME24 and AIME25 are released under Apache-2.0, MATH/MATH-500 and Minerva Math are released under MIT, and the official OlympiadBench dataset release is listed as Apache-2.0 while its accompanying official evaluation-code repository is MIT-licensed. The public AMC23 benchmark release used in our evaluation does not declare an explicit license on its current dataset card. The DeepScaleR evaluation framework is MIT-licensed.

For evaluation, we use the DeepScaler~\cite{deepscaler2025} evaluation framework with $32$ independent samples per prompt (avg@32), using temperature $1.0$ and $\text{top-}p=0.7$ at validation, matching the rollout configuration. All reported numbers are averages over $32$ decodes; random seeds for data shuffling and sampling follow the veRL defaults.

\section{Clipping the Adaptive Coefficient}
\label{app:clip}

A natural question is whether the online proxy $\lambda_t = \lambda_{+}\cdot\hat{c}_t/(\hat{v}_t^2+\epsilon)$ should be constrained to be positive $\max(0,\hat{c}_t)/\hat{v}_t^2$, that is, whenever the alignment proxy $\hat{c}_t$ becomes negative, should we (i) pass the negative value through (\textbf{OCC-NoClip}, our default), or (ii) clip it to zero (\textbf{OCC-Clip})?

\begin{table}[t]
\centering
\caption{Effect of clipping $\lambda_t$ to non-negative values. \textbf{OCC-Clip} clips the ratio to $\max(0,\hat{c}_t)/\hat{v}_t^2$; \textbf{OCC-NoClip} keeps the signed ratio. Final avg@32 accuracy (\%) on MiMo-7B-RL + DAPO. Best in \textbf{bold}.}
\label{tab:clip_ablation}
\resizebox{\linewidth}{!}{
\begin{tabular}{lcccccc|c}
\toprule
\textbf{Variant} & \textbf{AIME24} & \textbf{AIME25} & \textbf{AMC} & \textbf{MATH} & \textbf{Minerva} & \textbf{Olympiad} & \textbf{Avg.} \\
\midrule
OCC-Clip          & 43.9 & 44.1 & 81.8 & 90.3 & 42.1 & 59.6 & 60.3 \\
OCC-NoClip (Ours) & \textbf{45.9} & \textbf{46.7} & \textbf{83.1} & \textbf{91.0} & \textbf{43.4} & \textbf{60.3} & \textbf{61.7} \\
\bottomrule
\end{tabular}
}
\end{table}

Table~\ref{tab:clip_ablation} compares the two variants on MiMo-7B-RL + DAPO. OCC-NoClip outperforms OCC-Clip by $+1.3$ average points, with the largest gap on AIME25 ($+2.6$). We interpret this as follows. The optimal MTP coefficient at each step is proportional to the signed ratio $\lambda_t \propto \hat{c}_t / \hat{v}_t^2$, where $\hat{c}_t$ captures the alignment between RL and MTP gradients. 

Crucially, the closed-form optimum is symmetric in the sign of $c$: the maximum value of the parabola $\Delta_{\mathrm{MTP}}(\lambda)$ is $(1-L\eta)^2 c^2/(2Lv^2)\geq 0$ regardless of whether $c$ is positive or negative, and is attained at a $\lambda^*$ that shares the sign of $c$. When $c<0$, forcing $\lambda=0$ collapses the gain to zero, which is strictly worse than the negative-$\lambda^*$ optimum; clipping therefore replaces the maximizer with a strictly suboptimal projection on half of the parameter axis. Geometrically, a negative $\lambda$ flips the MTP gradient before it is added to the RL direction, so when the unflipped MTP gradient points against RL (i.e., $c<0$), the flipped version is by construction aligned with RL --- negative $\lambda$ is precisely the move that converts a hostile auxiliary signal into a corrective one. Clipping $\hat{c}_t$ to positive values (OCC-Clip) ignores these instances, preventing the model from applying corrective updates that could counteract harmful MTP directions. In contrast, keeping the signed ratio (OCC-NoClip) allows both positive and negative contributions to $\lambda_t$, enabling the model to dynamically adjust the MTP coefficient to maximize per-step improvement. This leads to more effective calibration throughout training, yielding consistently higher performance across tasks.

\section{Potential Risks}

Our work aims to improve the efficiency and stability of post-training large language models through joint MTP-RL optimization. While the method is designed to enhance reasoning, such improvements may also affect how models are used in broader settings. Since RLVR depends on verifiable rewards, models may become sensitive to the design and quality of the reward signal, making robustness checks important.

\section{Use of AI Assistants}

AI assistants were used solely to support the writing and editing process of this paper. Specifically, they were employed to polish the language, improve clarity and readability, and correct grammatical errors. All technical ideas, theoretical analysis, method design, experimental setup, results, and conclusions were developed and verified by the authors. The authors carefully reviewed and edited all AI-assisted text to ensure correctness, consistency, and alignment with the paper's intended meaning.

\end{document}